\documentclass[a4paper,10pt]{svproc}

\usepackage{type1cm}        
\usepackage{makeidx}         
\usepackage{graphicx}        
\usepackage{multicol}        
\usepackage[bottom]{footmisc}

\usepackage{graphicx}
\usepackage{xcolor}

\usepackage{newtxtext}       %
\usepackage[varvw]{newtxmath}       

\makeindex             
\usepackage{url}

\begin{document}
\mainmatter              
\title{Bootstrapping Robotic Skill Learning With Intuitive Teleoperation: Initial Feasibility Study}
\titlerunning{Bootstrapping Robotic Skill Learning With Intuitive Teleoperation}  
%
\author{Xiangyu Chu\inst{1,2,3,\dagger} \and Yunxi Tang\inst{1, \dagger}
Lam Him Kwok \inst{2} \and Yuanpei Cai\inst{4} \and \\ Kwok Wai Samuel Au\inst{1,2,4,*} }
\authorrunning{Xiangyu Chu and Yunxi Tang et al.} 
%
\tocauthor{Xiangyu Chu, Yunxi Tang,
Lam Him Kwok, Yuanpei Cai, Kwok Wai Samuel Au}
\institute{The Chinese University of Hong Kong, Hong Kong SAR\\
\and
Multi-Scale Medical Robotics Center, Hong Kong SAR
\and
Technical University of Munich, Germany
\and
Cornerstone Robotics Limited, Hong Kong SAR \\ $\dagger$: Equal contribution; *: samuelau@cuhk.edu.hk
}

\maketitle              

\begin{abstract}
Robotic skill learning has been increasingly studied but the demonstration collections are more challenging compared to collecting images/videos in computer vision and texts in natural language processing. This paper presents a skill learning paradigm by using intuitive teleoperation devices to generate high-quality human demonstrations efficiently for robotic skill learning in a data-driven manner. By using a reliable teleoperation interface, the da Vinci Research Kit (dVRK) master, a system called \textit{dVRK-Simulator-for-Demonstration} (dS4D) is proposed in this paper. Various manipulation tasks show the system's effectiveness and advantages in efficiency compared to other interfaces. Using the collected data for policy learning has been investigated, which verifies the initial feasibility. We believe the proposed paradigm can facilitate robot learning driven by high-quality demonstrations and efficiency while generating them.
\end{abstract}
\section{Introduction}
Motivated by the recent achievement of Large Language Models (LLMs), many researchers are interested in developing embodied intelligence and foundation models for autonomous robotic systems \cite{RT_X}. Although some promising results have been obtained in recent works (e.g., R3M \cite{R3M} and VC-1 \cite{VC1}), they are still far from working the best in all general tasks and fine tuning based on Pre-trained Visual Representations (PVRs) is needed. For example, VC-1 outperforms all prior PVRs on average but does not universally dominate. To improve skill learning results in specific downstream tasks, adaption with in-domain data (e.g., behaviour cloning on human demonstrations) is still necessary. Such a paradigm inspires us to bootstrap skill learning by seeking high-quality demonstrations to fine-tune pre-trained large models. Besides, the embodied AI community has developed many few-shot imitation learning algorithms, and they also require demonstrations for skill learning.

\begin{figure}[h]
\includegraphics[scale=.375]{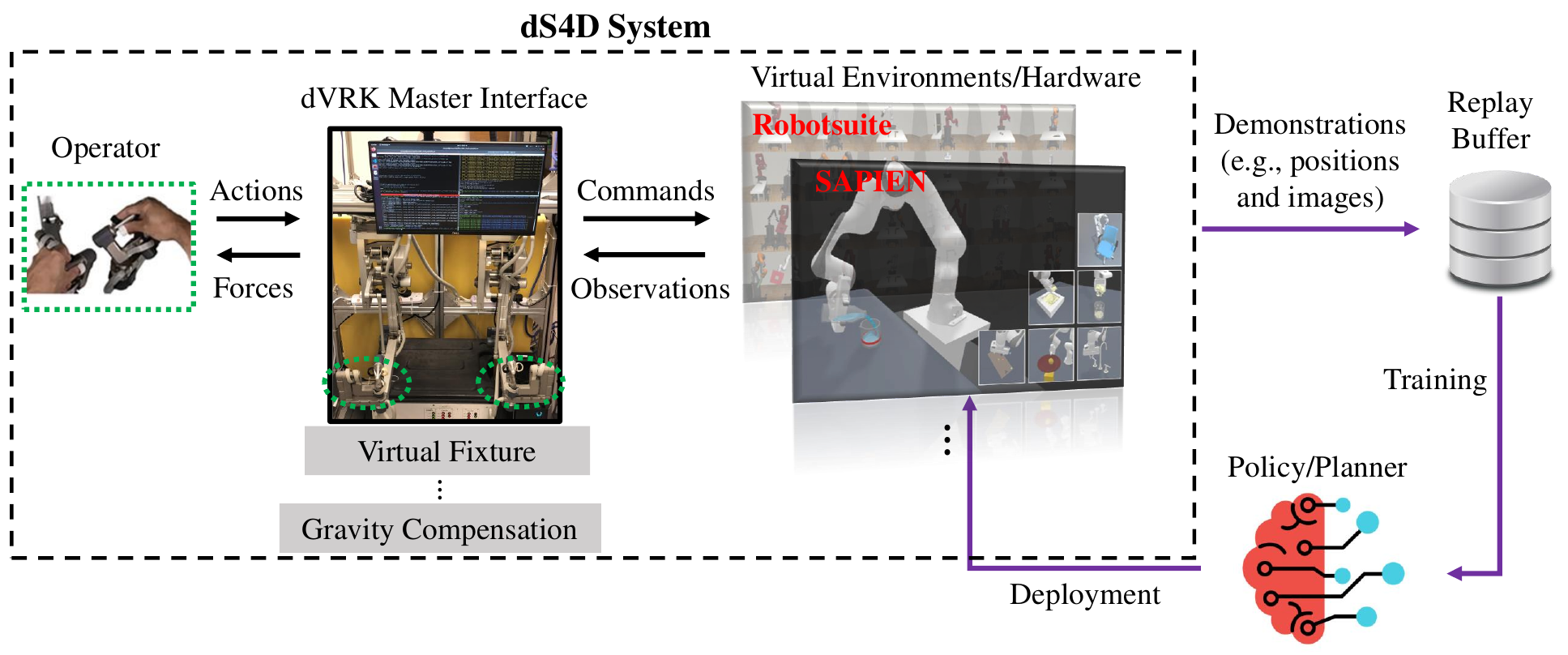}
%
%
\caption{Proposed skill learning framework based on the dS4D system. The operator can intuitively operate the robots in simulators by mapping the master's pose to that in simulators. The collected data is saved in a replay buffer and then fed to future policy learning.}
\label{framework}       
\end{figure}

To get demonstrations or collect data, except for scripted-based control and kinesthetic teaching, many interfaces have been used for connecting humans with robots, where the human operators make decisions and control movements; then the robots just execute and the execution information is logged. For example, keyboards can independently control each axis of translation and orientation \cite{keyboard}; 3D mouses/phones are essentially joysticks that can translate and rotate simultaneously \cite{Roboturk}; Virtual Reality (VR) controllers like Oculus Rift CV1 can map their pose to that of the robot's end-effector (EE) \cite{VR_controller}; VR headsets can also be used for hand-pose control in augmented reality \cite{Holo}. However, these interfaces suffer from time-consuming operations while conducting tasks.
In this paper, we will focus on the master interface of a da Vinci Research Kit (dVRK) system, which has been widely used in robotic surgeries due to its intuitiveness and high precision. 
Although the dVRK master has been used to teleoperate bimanual arms to generate human demonstrations \cite{dVRK_2017}, no quantitative comparisons with other interfaces have been conducted. Besides high data collection efficiency, we will also show that the collected data is high-quality such that the learned policy can achieve comparable performance in evaluation.


In this work, we build a system named dVRK-Simulator-for-Demonstration (dS4D), which can be utilized to connect the dVRK master with various simulators (and robotic hardwares) in different environments such as general manipulation tasks \cite{robotsuite}, housekeeping tasks \cite{iGibson2}, and deformable object manipulation tasks \cite{SAPIEN}. Within simulators, we are able to obtain a range of observations from proprioceptive and exteroceptive sensors, which are suitable for data-driven methods in robot skill programming. This paper will choose a simulator, \texttt{RobotSuite}, for a feasibility study.

\section{Related Work}
\textbf{Demonstration Collection:} To enhance robotic skill acquisition, an effective way is to learn complex skills via human demonstrations. There are three main types of interfaces that allow humans to generate demonstrations: kinesthetic interfaces, customized interfaces, and teleoperation interfaces. At the early stage of learning from demonstration, kinesthetic interfaces were widely used because humans could directly interact with the robots without involving extra devices \cite{TableTennis,mime}, but they may result in occlusion in the collected images and collecting large-scale data based on them is challenging. To collect the demonstration at scale and exclude humans' occlusion, researchers built customized interfaces using plastic grabber tools equipped with cameras and necessary electronic components \cite{PlasticGraspper,PlasticGraspper2}. Although users can operate customized interfaces for intuitive grasping at low cost, additional processing like extracting gripper trajectory and state is required. Instead of interacting with robots/tools directly, teleoperation interfaces, such as 3D motion controller \cite{Visuomotor2018Zhu}, VR \cite{Holo}, and bimanual devices \cite{ALOHA}, allow users to send control commands in a remote setting. More importantly, they can help collect versatile observations (e.g., the robot's states and the external sensor's visual information) while connected to various simulators and hardware. In this paper, we choose to use a reliable teleoperation interface, the dVRK master, due to its intuitiveness and efficient demonstration collection.


\noindent \textbf{Skill Learning with Demonstrations:} Imitation learning is one of the main approaches for skill learning from demonstrations and a policy is trained from the demonstrations. As the simplest imitation learning method, Behavior Cloning (BC) learns a map from states to actions from the demonstrations via supervised learning \cite{Alvinn}. There are several variants of BC such as BC-RNN \cite{robotmimic} and Hierarchical BC (HBC) \cite{GTI} \cite{CollaborativeTeleoperation2021}. In BC-RNN, a Recurrent Neural Network (RNN) is used to serve as a policy network, integrating temporal dependencies into the policy via recurrent hidden states. Specifically, a sequence of data, such as $(s_t, a_t, ..., s_{t+T}, a_{t+T})$, is used in training the network, where $s_t$ is the state at $t$-th step and $a_t$ is the action. HBC has two layers: 1) a low-level policy is conditioned on future observations and outputs a sequence of actions and 2) a high-level policy outputs the future observations based on the current observations. Besides offline imitation learning methods, batch (Offline) reinforcement learning, such as Batch Constrained Q-Learning (BCQ) \cite{BCQ} and Conservative Q-Learning (CQL) \cite{CQL}, can learn policies from the demonstrations, especially from suboptimal data. To study the quality of the collected data, we choose to use both offline imitation learning and reinforcement learning methods for skill learning.

\section{Method}
In this section, we describe the dVRK-Simulator-for-Demonstration (dS4D) framework for collecting human demonstrations and the study of a teleoperation protocol.

\subsection{Proposed Framework}
A skill learning framework based on the dS4D system is proposed, as shown in Fig. \ref{framework}, where the task demonstrations generated by the dS4D system can be used for policy learning from human demonstrations via behavior cloning and offline reinforcement learning approaches.

The input device is the dVRK master (in the left of Fig. \ref{framework}) with two 6 Degree-of-Freedom (DoF) Masters Tool Manipulators (MTMs). Each MTM has a two-finger gripper and provides grasping with controllable width. If needed, foot pedals can be used for complementary functions such as mode switch, start/stop/end, and clutch. Although the depth is missed when the operator is observing a monitor, extra visualization like the pointing direction of the robot EE can help recognize the depth. 
We map the pose of the master's EE to that in the simulator, instead of mapping the joint angles, to provide intuition to operators. 
A block of gravity compensation \cite{GC_ben} is included to reduce muscle fatigue caused by lifting the weight of the master devices. Given a task, the operator can use the master to generate demonstrations, which will be saved in a buffer and then used for training policies after the necessary post-processing of data.


The dS4D system is mainly built on \texttt{ROS}, which is the bridge for communication between the master device and simulator environments. On the MTM side, an open-source software architecture \cite{dvrk} is used for all levels of MTM control and communications. The simulator runs as a ROS node, i.e., \texttt{SimulatorNode}, that subscribes the teleoperation commands from the \texttt{InterfaceNode} and publishes the states of the simulated environment (such as the robot joint states and images) for planners/controllers and visualization utils. The \texttt{InterfaceNode} connects the simulators and MTM device hardware, and subscribes to sensor topics from the MTM device, such as the MTM tip pose, gripper states, and foot-pedal states. With the obtained information from the master device, \texttt{InterfaceNode} post-processes the inputs of the master device and publishes the action commands to the robot arm controllers in the simulator. Technically, various interface protocols can be implemented in \texttt{InterfaceNode} for specific tasks. Hence, the teleoperation settings for demonstration collection can be flexibly customized for different simulator environments and manipulation tasks in a decoupled manner.

\subsection{Teleoperation Protocol Example}
In this study, we use a simple teleoperation protocol as an illustrative example in the \texttt{InterfaceNode} to show how dS4D can be utilized for efficient demonstration collection. 
To be specific, the \texttt{InterfaceNode} translates the pose/transform and the gripper finger position of the MTM tip into the corresponding counterparts of the robots in the simulator. 

We firstly give some fundamental monogram notations. The transform matrix of frame $\{A\}$ measured from frame $\{B\}$ is denoted as 
\begin{equation}
    {^A}\mathbf{X}_B = \left[
    \begin{array}{cc}
        {^A}\mathbf{R}_B & {^A}\mathbf{p}_B \\
         \mathbf{0} & 1
    \end{array}
    \right] 
    \in SE(3).
\end{equation}
The transform ${^A}\mathbf{X}_B$ includes translation $\mathbf{p} \in \mathbb{R}^3$ and rotation $\mathbf{R} \in SO(3)$. The following frames and transforms are used in the teleoperation protocol example:
\begin{itemize}
    \item $\{B\}$: Base frame of the MTM device. This fixed frame serves as the reference for the poses published by the dVRK master;

    \item $\{R\}$: Reference frame of the operator, which is assumed to be a constant during teleoperation. ${^R}\mathbf{X}_{B}$ is known for a teleoperation setting;

    \item $\{M\}$: MTM tip frame. ${^B}{\mathbf{X}}_{M}$ is the MTM tip transform, which can be calculated with the forward kinematics of the master;

    \item $\{W\}$: World frame of the robot arm(s) in the simulator. This fixed frame is referenced by the simulator;

    \item $\{C\}$: Camera frame in the simulator. This fixed camera is the default one that used for the monitor display and ${^C}\mathbf{X}_{W}$ is known;
    
    \item $\{G\}$: Robot gripper frame in the simulator. ${^W}{\mathbf{X}}_{G}$ is the transform of the robot EE in the simulator, which can be obtained with the forward kinematics of the robot.
\end{itemize}

During demonstration collection, the transform ${^C}\mathbf{X}_{G}$ and ${^R}\mathbf{X}_{M}$ should be consistently aligned for intuitive teleoperation. Hence, the pose mapping at every time step is given as below
\begin{equation}
\begin{aligned}
        \text{desired translation: }& {^C}\mathbf{p}^d_{G} = {^C}\mathbf{p}_{W}+ {^C}\mathbf{R}_{W}{^W}\mathbf{p}^0_{G} + \lambda \; {^R}\mathbf{R}_{B}({^B}\mathbf{p}_M - {^B}\mathbf{p}^0_M), \\
        \text{desired orientation: }& {^C}\mathbf{R}^d_{G} = {^R}\mathbf{R}_{B}\; {^B}\mathbf{R}_{M}.
\end{aligned}
\end{equation}
where $\lambda>0$ is a scale factor. ${^W}\mathbf{p}^0_{G}$ and ${^B}\mathbf{p}^0_M$ are the initial positions of the robot gripper and MTM tip, respectively. Note that the robot is controlled to adjust its initial gripper orientation ${^W}\mathbf{R}^0_{G}$ to align with ${^R}\mathbf{R}^0_{M}$ at the beginning. Since the robot is controlled by an operational space controller (OSC), we represent the desired transform of $\{G\}$ with respect to the simulator world frame $\{W\}$ as below
\begin{equation}
    {^W}\mathbf{X}^d_G = {^W}\mathbf{X}_C \left[
    \begin{array}{cc}
        {^C}\mathbf{R}^d_G & {^C}\mathbf{p}^d_G \\
         \mathbf{0} & 1
    \end{array}
    \right],
\end{equation}
then ${^W}\mathbf{X}^d_G$ can be tracked by the low-level OSC controller. For the grasping motion, the width of gripper is directly controlled by the gripper of MTM.






\begin{figure}[h]
\includegraphics[scale=.5]{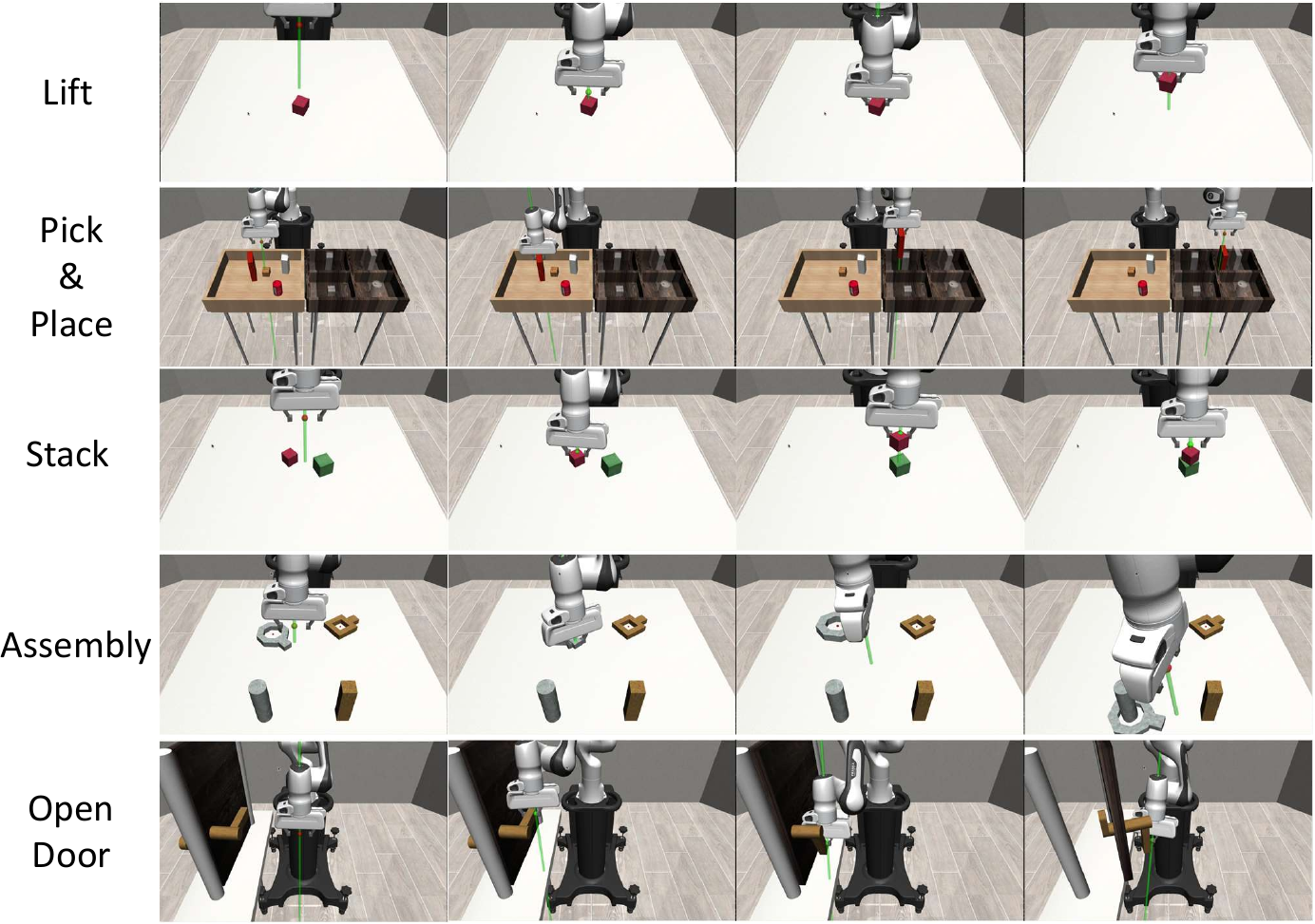}
%
%
\caption{Five tasks implemented by the dS4D system.}
\label{snapshot_robotsuite}       
\end{figure}

\begin{figure}[]
\includegraphics[scale=.5]{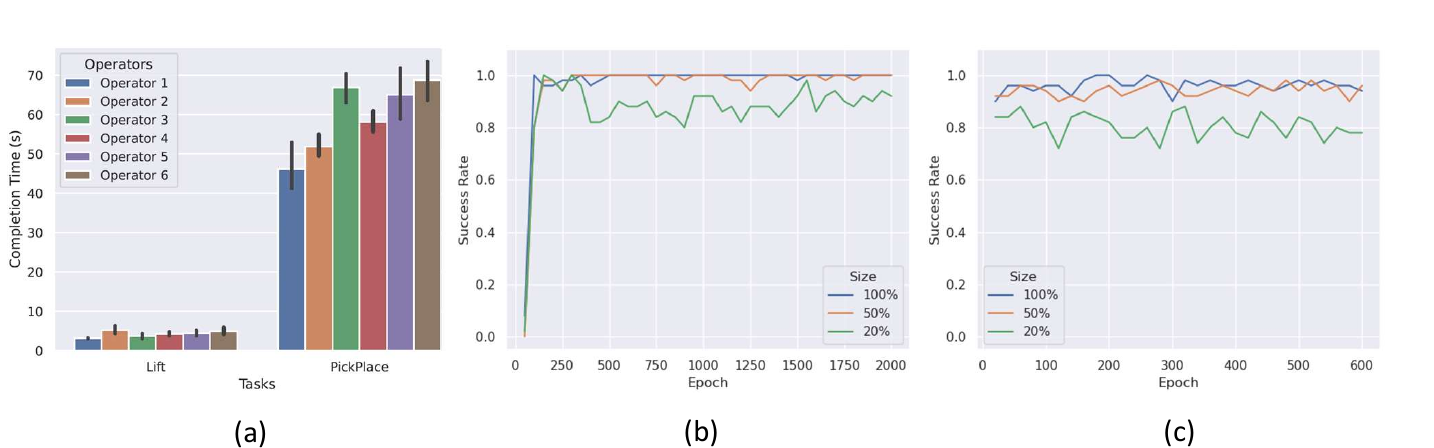}
%
%
\caption{(a) Completion time of two tasks conducted by six operators. (b-c) Success rate for the Lift and Pick-and-Place tasks, respectively.}
\label{WithoutVF}       
\end{figure}

\section{Results}
We conducted experiments to illustrate the intuitiveness and effectiveness of the dVRK master in collecting demonstration data. All the experiments were completed based on the teleoperation protocol in the previous section. The collected demonstrations were used for skill learning and comparable even better performance could be observed.


\begin{table}[!]
\caption{Completion time statistics of the lift (\textit{pick-and-place}) task using different user interfaces}
\label{Statistics}       
%
%
\begin{tabular}{p{3cm}p{2.4cm}p{3cm}p{2.6cm}}
\hline\noalign{\smallskip}
Devices & Mean (s) & Standard Deviation (s) & No. Demonstrations  \\
Keyboard \cite{Roboturk} & 22.34 (\textit{151.45})  & 7.92 (\textit{23.69}) & 187 (\textit{40}) \\
3D Mouse \cite{Roboturk} & 16.58 (\textit{112.57}) & 12.27 (\textit{23.69}) & 193 (\textit{40})\\
VR Controller \cite{Roboturk} & 14.29 (\textit{79.36})  & 8.38 (\textit{29.41})& 160 (\textit{40})\\
Phone \cite{Roboturk} & 14.26 (\textit{89.97})  & 8.27 (\textit{34.94}) & 167 (\textit{40})\\
dVRK Master & \textbf{3.54} (\textbf{\textit{57.73}}) & \textbf{1\textit{.28}} (\textbf{\textit{13.31}}) &238 (\textit{70}) \\ 
\noalign{\smallskip}\hline\noalign{\smallskip}
\end{tabular}
\end{table}

\subsection{Data Collection}
We firstly show how dS4D can be used to collect human demonstrations for robot skill learning. A commonly-used robotic simulation environment, \texttt{Robosuite} \cite{robotsuite}, was connected with the dVRK master device. Specifically, the master was connected to a Panda manipulator, in which the EE's pose and gripper's finger position can be teleoperated by MTM, and the operator was expected to use the MTM device to implement different types of manipulation tasks based on visual feedback (via monitor). Five tasks were conducted to show the effectiveness of the dS4D system (Fig. \ref{snapshot_robotsuite}) in demonstration generation. In the Lift task, a small cube was expected to be lifted. In the Pick-and-Place task, a Coke can was expected to be moved to a target bin. In the Stack task, a red cube was expected to be picked up and placed over a green cube.  In the Assembly task, a nut was expected to be picked and placed into a rod. In the Open Door task, the door was expected to be opened by grasping the door handle.  For comparison with other interface devices, the Lift and Pick-and-Place tasks were conducted by six operators.


In Table 1, the dVRK master is demonstrated to outperform all other interfaces in terms of completion time, which implies the dVRK master is more efficient and intuitive than other devices. Interestingly, we found that the operators with different proficiency could finish the manipulation tasks within a similar time duration (low standard deviations, see Fig. \ref{WithoutVF} (a)) and all their consumed time on average is less than that using other devices. Moreover, even fresh operators can reach a good teleoperation performance without extensive pre-training. The intuitiveness and usability of the master are useful for generating large demonstration datasets for robot skill learning in a reliable and fast way. 



\begin{figure}[]
\includegraphics[scale=.6]{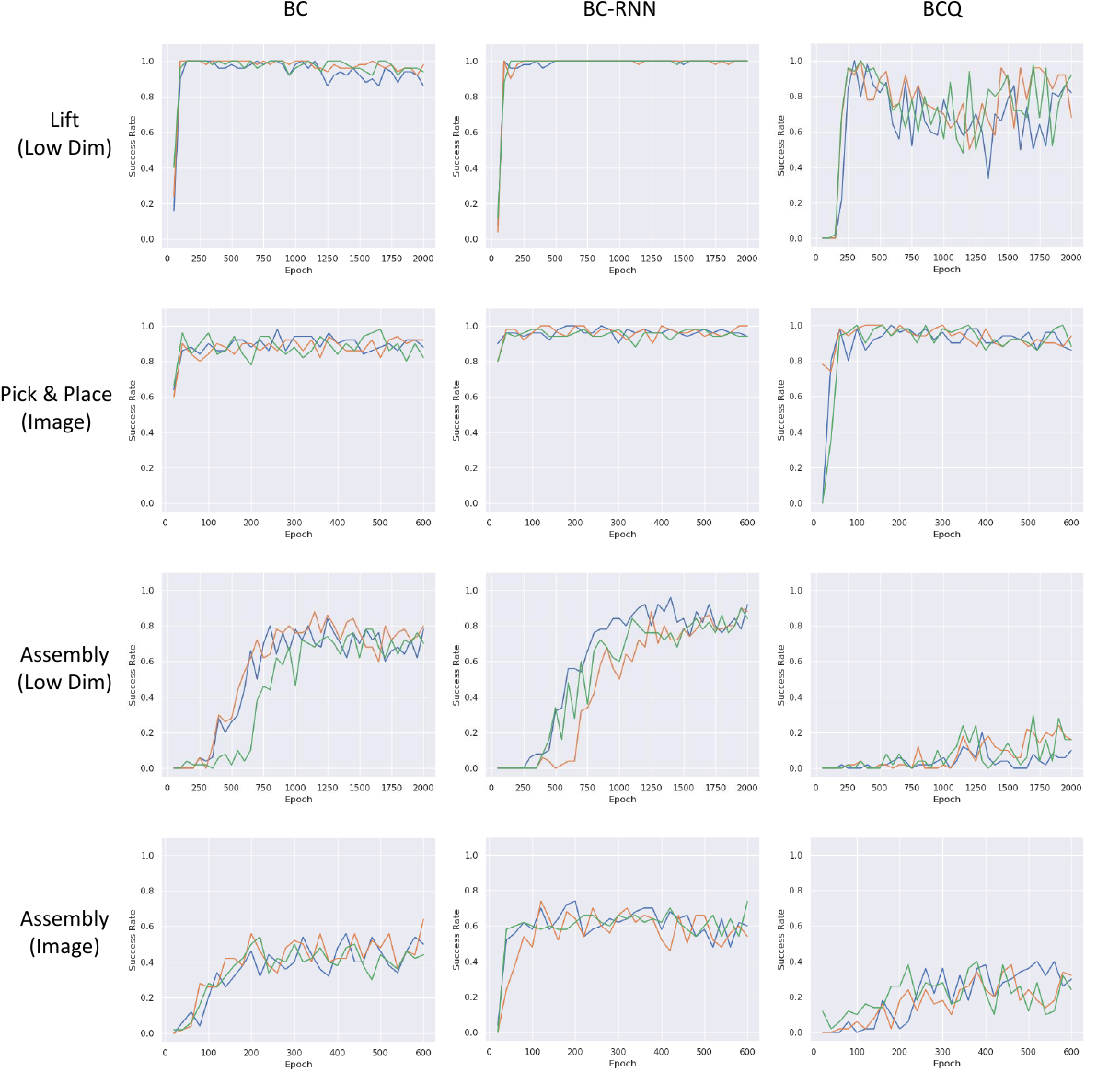}
%
%
\caption{Learning curves of using four datasets for three algorithms.}
\label{learning_curves}       
\end{figure}

\begin{table}[!]
\caption{Average success rates across 3 seeds}
\label{success_rate}       
%
%
\begin{tabular}{p{3cm}p{2.4cm}p{3cm}p{2.6cm}}
\hline\noalign{\smallskip}
Dataset & BC & BC-RNN & BCQ  \\
Lift (Low Dim) &  $\textbf{100.0} \pm \textbf{0.0}$  & $\textbf{100.0} \pm \textbf{0.0}$ & $\textbf{100.0} \pm \textbf{0.0}$ \\
Pick-and-Place (Image) & $96.6 \pm 2.3$ & $\textbf{99.3} \pm \textbf{1.1}$ & $\textbf{100.0} \pm \textbf{0.0}$\\
Assembly (Low Dim) & $83.2 \pm 5.0$  & $\textbf{92.0} \pm \textbf{3.5}$ & $24.3 \pm 5.0$\\
Assembly (Image) & $57.8 \pm 5.3$  & $\textbf{74.0} \pm \textbf{0.0}$ & $39.3 \pm 1.2$\\
\noalign{\smallskip}\hline\noalign{\smallskip}
\end{tabular}
\end{table}

\subsection{Training and Evaluation}
Offline imitation learning or offline reinforcement learning approaches can be used in robot skill learning. In this paper, we employed two offline imitation learning methods (BC and BC-RNN) and one offline reinforcement learning (BCQ) to imitate the manipulation skills from the human demonstrations generated by dS4D. Three tasks were studied, including Lift, Pick-and-Place, and Assembly \cite{robotmimic}. We trained and evaluated three algorithms on the multi-operator datasets. ``Low Dim" means the object observation consists of the object's absolute pose and relative pose to the robot EE. ``Image" means the object observation consists of the images from external and wrist cameras. Both ``Low Dim" and ``Image" consist of the robot EE's pose and gripper's position.

The average success rates of the three seeds are shown in Table 2 and the learning curves are shown in Fig. \ref{learning_curves}. With the demonstration dataset collected by dS4D, a less complex task, i.e., Lift, can achieve $100$\% success rate during all tests. Complex tasks like Pick-and-Place and Assembly using BC-RNN can also achieve $99.3$\% and $92.0\%$ success rates, respectively, which is comparable to the performance of \cite{robotmimic} in the Pick-and-Place task and even better than the performance of \cite{robotmimic} in the Assembly task. We found that, in the realm of offline imitation learning, introducing temporal dependencies greatly improves the performance; for example, the BC-RNN agent outperformed the BC agent in the Assembly task with either low-dim observations (92.0\% vs. 83.2\%) or image observations (74.0\% vs. 57.8\%). The offline reinforcement learning method, BCQ, performed well in the tasks of Lift and Pick-and-Place. However, it performed poorly in the Assembly task and the success rate was less than 40\%. The image observations were found to be effective in learning policies from high-dimensional observations, but the low-dim agents still outperformed the image agent in the Assembly task. Lastly, we studied the effect of the dataset size on the performance and created two subsets, 20\% and 50\% of the dataset. In Fig. 3 (b-c), the results show that, in the Lift and Pick-and-Place tasks, using a part of the dataset can still achieve superior performance. Specifically, the low-dim lift agent can still achieve 100\% success rates with 20\% or 50\% of the dataset. In the Pick-and-Place task, we found increasing the dataset size can improve the performance, which emphasizes the value of large human demonstrations.




\section{Main Experimental Insights}
Our experiments demonstrate the feasibility of the proposed skill learning paradigm. The intuitive dVRK teleoperation device can be connected to various simulators, and such flexibility allows users to collect demonstration data either in the off-the-shelf simulators or in customized virtual environments. Due to the intuitiveness of our dS4D system, users are allowed to collect data more efficiently. Although many interfaces have been used for demonstration, the dS4D system can complete the same demonstration by using much less time. Such an advantage in efficiency also raises awareness of improving efficiency in data collection because people normally collect data at any cost. Moreover, the policies trained based on our data show comparable and even better performance than the data from Phones. This implies the data generated by the dS4D system still keeps high quality even though the collection is fast.

Overall, our experiments reinforce the quantitative analysis of various teleoperation interfaces and emphasize the benefits of the dVRK master in demonstration collection, facilitating the development of robotic skill learning in more complex tasks and more challenging fields. In the future, we will further verify the feasibility of the proposed paradigm in more complex tasks such as deformable object manipulation tasks in the customized environment based on MuJoCo. Besides, connecting the dVRK master with hardware like robotic arms for manipulation tasks is also promising. In \cite{RT_X}, a high-capacity model, RT-X, for robotic manipulation benefited from a large-scale dataset where more than 50\%  of the dataset relied on teleoperation, which further inspires us to use our method to contribute to the data collection with higher efficiency.
%
%

\end{document}